\documentclass[10pt,twocolumn,letterpaper]{article}

\usepackage[pagenumbers]{cvpr} 
\usepackage{algpseudocode}
\usepackage{algorithm}
\usepackage{stfloats}
\usepackage{cuted}
\usepackage{multirow}
\usepackage{color}
\usepackage{colortbl}
\definecolor{cvprblue}{rgb}{0.21,0.49,0.74}
\usepackage[pagebackref,breaklinks,colorlinks,allcolors=cvprblue]{hyperref}

\def\paperID{7004} 
\def\confName{CVPR}
\def\confYear{2025}

\title{Scene Splatter: Momentum 3D Scene Generation from Single Image \\ with Video Diffusion Model}

\author{Shengjun Zhang$^{1}$, Jinzhao Li$^{1}$, Xin Fei$^{1}$, Hao Liu$^{2}$, Yueqi Duan$^{1\dag}$\\
$^{1}$Tsinghua University, $^{2}$WeChat Vision, Tecent Inc. \\
{\tt\small \{zhangsj23, lijinzha22\}@mails.tsinghua.edu.cn, duanyueqi@tsinghua.edu.cn}}

\begin{document}
\maketitle
\newcommand\blfootnote[1]{%
\begingroup 
\renewcommand\thefootnote{}\footnote{#1}%
\addtocounter{footnote}{-1}%
\endgroup 
}
\blfootnote{\textsuperscript{\dag}Corresponding author.}
\begin{strip}
    \vspace{-2.3cm}
    \begin{center}
    \textbf{\url{https://shengjun-zhang.github.io/SceneSplatter/}}
    \end{center}
    \centering
    \includegraphics[width=\linewidth]{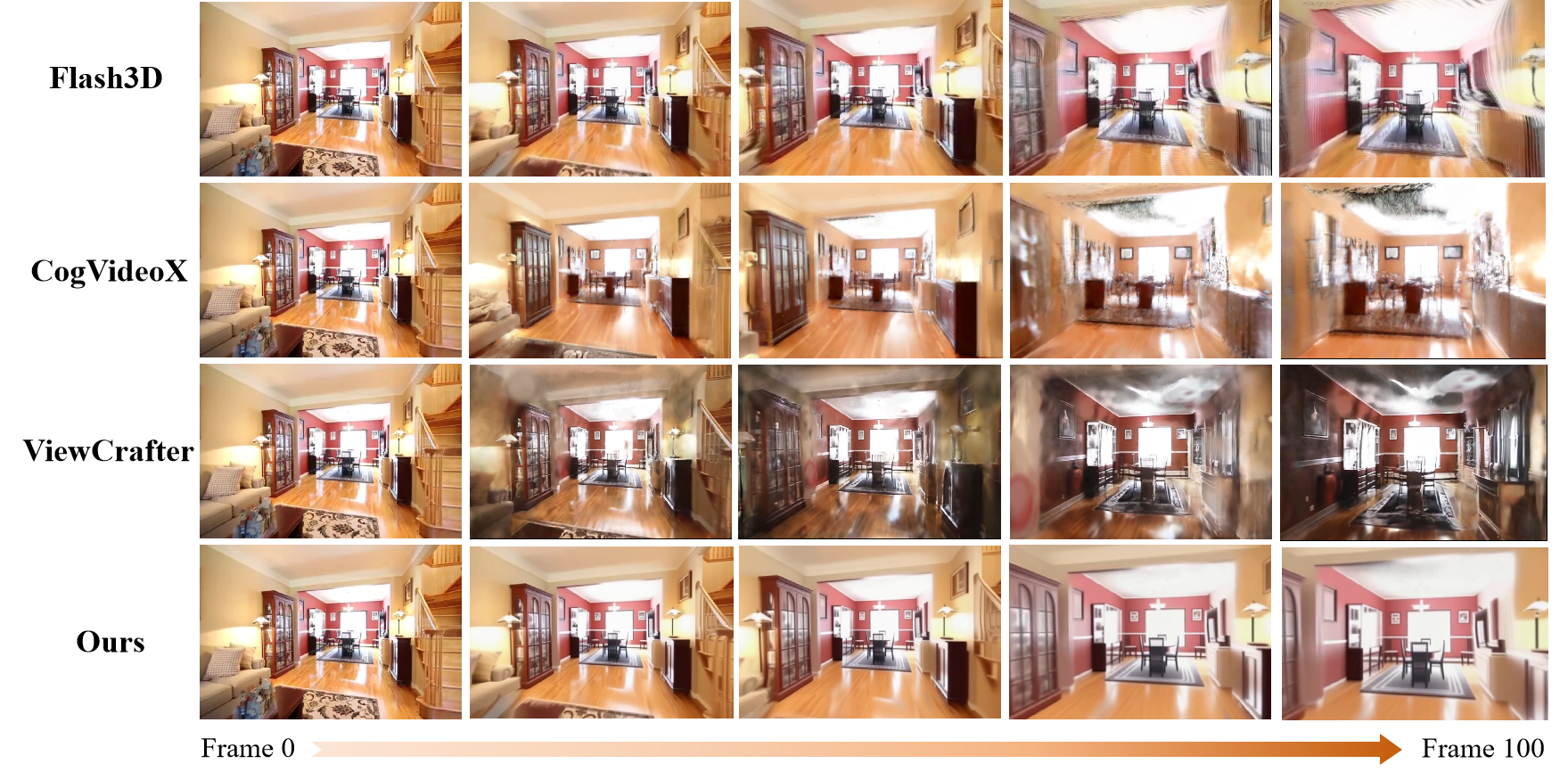}
    \captionof{figure}{Visualization results of Flash3D~\cite{Flash3D2024NIPS}, CogVideoX~\cite{CogVideoX2024arXiv}, ViewCrafter~\cite{ViewCrafter2024arXiv} and ours. Flash3D~\cite{Flash3D2024NIPS} suffers from distortions and occlusions, while CogVideoX~\cite{CogVideoX2024arXiv} and ViewCrafter~\cite{ViewCrafter2024arXiv} change the color style or existing components, compared to the input image. Our method can generate high fidelity and consistent 3D scene with our cascaded momentum.}
    \label{fig:teaser}
\end{strip}

\begin{abstract}

In this paper, we propose Scene Splatter, a momentum-based paradigm for video diffusion to generate generic scenes from single image.
Existing methods, which employ video generation models to synthesize novel views, suffer from limited video length and scene inconsistency, leading to artifacts and distortions during further reconstruction.
To address this issue, we construct noisy samples from original features as momentum to enhance video details and maintain scene consistency.
However, for latent features with the perception field that spans both known and unknown regions, such latent-level momentum restricts the generative ability of video diffusion in unknown regions.
Therefore, we further introduce the aforementioned consistent video as a pixel-level momentum to a directly generated video without momentum for better recovery of unseen regions.
Our cascaded momentum enables video diffusion models to generate both high-fidelity and consistent novel views.
We further finetune the global Gaussian representations with enhanced frames and render new frames for momentum update in the next step.
In this manner, we can iteratively recover a 3D scene, avoiding the limitation of video length.
Extensive experiments demonstrate the generalization capability and superior performance of our method in high-fidelity and consistent scene generation.

\end{abstract}    
\section{Introduction}
Recovering 3D models from 2D images is a fundamental problem in computer vision, due to its widespread applications, such as virtual reality, augmented reality, robotics and so on.
Recently, 3D neural reconstruction techniques, such as NeRF~\cite{NeRF2021ACM} and 3D Gaussian Splatting (3DGS)~\cite{3DGS2023ToG} has achieved remarkable progress.
Despite of their high quality in dense view reconstruction, generating a scene from one single view is still an ill-posed problem, since unambiguous geometric cues and occluded parts are unavailable in the monocular setting.

Previous methods attempt to learn scene prior knowledge from additional training data with various techniques, such as ResNet~\cite{pixelNeRF2021CVPR, Flash3D2024NIPS}, epipolar transformer~\cite{pixelSplat2024CVPR} and cost volume~\cite{MVSNeRF2021ICCV, MVSplat2024ECCV}, to synthesize novel views from sparse or single input.
However, these methods struggle to acquire high-quality renderings in unseen areas, especially with out-of-distribution input data.
As shown in Figure~\ref{fig:teaser}, Flash3D~\cite{Flash3D2024NIPS} fails to recover unknown regions and suffers from distortions in geometry.
Recent advancements in powerful generative models have promoted 3D generation from single input. 
Some methods~\cite{Zero1232023ICCV, MVdream2023arXiv, Instant3d2023arXiv, Sweetdreamer2023arXiv} generate multi-view images for reconstruction, but they are restricted to object-level generation. 
Other methods~\cite{ScenScape2023, RealmDreamer2024arXiv, VividDream2024arXiv} apply 2D diffusion models for per-view inpainting.
Yet, they often generate inconsistent contents with semantic drifts.

We analogy novel view synthesis to the generation of continuous video frames,
and reformulate the challenging 3D consistency problem as temporal consistency within video generation, to unleash the strong generative prior of pre-trained large video diffusion models~\cite{3DGS-Enhancer2024NeurIPS, ViewCrafter2024arXiv}.
An intuitive idea is to directly recover 3D scene from generated videos.
As illustrated in Figure~\ref{fig:teaser}, we adopt CogVideoX~\cite{CogVideoX2024arXiv} and ViewCrafter~\cite{ViewCrafter2024arXiv} to enhance the view synthesis process, but they change the property of existing components, leading to conflicts in 3D reconstruction.
To address these issues, we propose Scene Splatter, a momentum 3D scene generation paradigm to introduce existing scene information as momentum in the generation process, to balance the generative prior and scene consistency.
Specifically, we construct noisy samples from original features as latent-level momentum to guide each denoising step, avoiding the change of existing components in multiple steps of reverse diffusion.
For latent features covering both known and unknown regions, latent-level momentum restricts the generation of unseen regions.
Therefore, we decode the denoised latent features to RGB space and inject this consistent video as pixel-level momentum to a generated video without aforementioned latent-level momentum for further enhancement of unknown areas. 
We finetune the global Gaussian representations supervised by the enhanced views. 
Instead of high-level camera pose prompts~\cite{CameraCtrl2024arXiv, MotionCtrl2024SIGGRAPH}, we follow our predefined camera trajectory to obtain rendering results from Gaussian representations for momentum update in the next generation step.
In this manner, we can iteratively recover a 3D model from one single image, avoiding the inherent restriction of video length in diffusion models.

Experimental results show that our method outperforms regression-based and generation-based baselines for high quality and scene consistency, unveiling the immense potential to create complex 3D scenes using video diffusion models.
Our contributions can be summarized as follows:
\begin{enumerate}
    \item We introduce Scene Splatter, a momentum based framework that generates 3D scenes from one single image while maintaining high quaility and scene consistency.
    \item We construct latent-level and pixel-level momentum from Gaussian representations for generation process to balance generative prior and existing scene information.
    \item Qualitative and quantitative results demonstrate that our method achieve superior performance  with high-fidelity and generalizability.
\end{enumerate}

\section{Related Works}
\label{sec:related works}

\subsection{Regression Model for 3D Reconstruction}
Recent advancements in neural representations have promoted the development of 3D reconstruction.
Neural Radiance Field~\cite{NeRF2021ACM} and 3D Gaussian Splatting~\cite{3DGS2023ToG} require dense input views for per-scene optimization, and suffer from overfitting with few inputs.
Early attempts~\cite{pixelNeRF2021CVPR, MVSNeRF2021ICCV, Grf2021ICCV, Sharf2021CVPR, SRF2021CVPR} on NeRF opt for pretraining on extensive datasets to impart prior knowledge.
Similar ideas~\cite{pixelSplat2024CVPR, MVSplat2024ECCV, Flash3D2024NIPS, FreeSplat2024NIPS, Splatterimage2024CVPR, GGN2024NIPS} are adopted to 3DGS, to predict scene-level Gaussian representations.
For sparse view inputs, pixelSplat~\cite{pixelSplat2024CVPR} applies an epipolar transformer to extract scene features from image pairs, while MVsplat~\cite{MVSplat2024ECCV} constructs cost volume representation via plane sweeping to produce 3D Gaussians in a faster way.
For single view inputs, Flash3D~\cite{Flash3D2024NIPS} adopts a pre-trained depth network and predict multi-layer Gaussians.
However, they are limited by the scarcity and diversity of 3D data and struggle to acquire accurate geometric cues in unseen areas for high-quality renderings .

\subsection{Generative Model for 3D Reconstruction}
The rapid development in 2D diffusion models~\cite{DDPM2020NIPS, DDIM2021ICLR, LDM2022CVPR} have shown exceptional generative capability, which can be leveraged for 3D Generation.
Previous studies distill the knowledge in the pre-trained 2D diffusion models~\cite{HierarchicalT2V2022, PhotorealisticT2I2022NIPS} into a coherent 3D model with the Score Distillation Sampling technique~\cite{Reconfusion2024CVPR, Magic3d2023CVPR, Prolificdreamer2024NIPS}.
To enhance the 3D consistency, several works~\cite{Zero1232023ICCV, Wonder3D2024CVPR, One23452023NeurIPS, LGM2024ECCV} have conditioned 2D diffusion models on multi-view camera poses.
For example, Wonder3D~\cite{Wonder3D2024CVPR} employs a multi-view cross-domain attention mechanism to exchange information across views and modalities.
Yet, these methods are restricted to object-level generation.
Some other researches~\cite{RealmDreamer2024arXiv, WonderJourney2024CVPR} follow the iterative processes of unprojecting, rendering and outpainting to synthesize novel views from a single image. 
Nevertheless, they often suffer from inconsistent generated contents and semantic drifts. 
More recently, video diffusion models~\cite{LVDM2022arXiv, SVD2023arXiv, Dynamicrafter2024ECCV, Videocrafter2024arXiv} have shown an impressive ability to produce realistic videos for 3D generation~\cite{V3D2024arXiv, SV3D2024ECCV}.
However, some methods~\cite{ViewCrafter2024arXiv, CogVideoX2024arXiv} change the property of existing components when enhancing input videos, while other methods~\cite{CameraCtrl2024arXiv, MotionCtrl2024SIGGRAPH} lack precise control of the camera motion due to their explicit injection of high-level camera pose prompts to video diffusion models.



\section{Methods}

\begin{figure*}
    \centering
    \includegraphics[width=\linewidth]{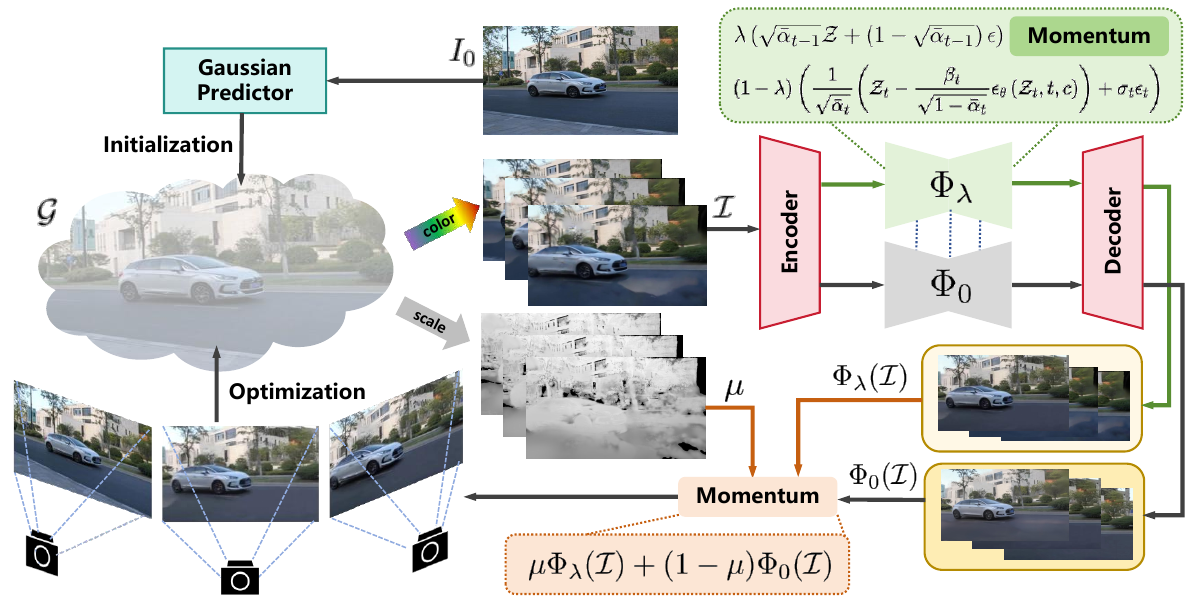}
    \caption{The pipeline of Scene Splatter. We initialize the Gaussian representations from the input image $I_{0}$ with a Gaussian Predictor~\cite{Flash3D2024NIPS}. For each iteration, we first render the video $\mathcal{I}$ from 3D Gaussians $\mathcal{G}$. Then, we generate the enhanced video $\Phi_{\lambda}(\mathcal{I})$ with latent-level momentum and $\Phi_{0}(\mathcal{I})$ directly from the vanilla diffusion model, where $\Phi_{\lambda}$ and $\Phi_{0}$ share the same weights of the denoising network. We further render scale maps as pixel-level momentum coefficient to further enhance the generated frames. We use the final results to supervise the optimization of Gaussian representations. We conduct this process along the camera trajectory to iteratively recover 3D scenes.}
    
    \label{fig:pipeline}
\end{figure*}

The pipeline of our framework is illustrated in Figure~\ref{fig:pipeline}.
In Section~\ref{sec:preliminary}, we briefly review 3D Gaussian Splatting and video diffusion models. 
In Section~\ref{sec:momentum scene generation}, we propose our latent-level and pixel-level momentum for scene generation.
In Section~\ref{sec:overall architecture}, we introduce details of the overall architecture to recovery 3D scenes from single inputs.

\subsection{Preliminary}\label{sec:preliminary}

\textbf{3D Gaussian Splatting.}
3DGS~\cite{3DGS2023ToG} represents a scene as a set of 3D Gaussian primitives, including a center position $\mu\in\mathbb{R}^{3}$, a covariance matrix $\Sigma\in\mathbb{R}^{3\times 3}$, an opacity $\alpha\in[0,1)$ and spherical harmonics coefficient $c\in\mathbb{R}^{k}$.
The Gaussian function can be formulated as:
\begin{equation}
    G(x) = e^{-\frac{1}{2}(x-\mu)^\top\Sigma^{-1}(x-\mu)}, \label{eq: Gaussian function}
\end{equation}
where $\Sigma=RSS^\top R^\top$, $S$ is the scaling matrix and $R$ is the rotation matrix.
For every pixel, the color is rendered by a set of Gaussians sorted in depth order:
\begin{equation}
    C=\sum_{i\in N} c_{i}\alpha_{i}\prod_{j=1}^{i-1}(1-\alpha_{i}). \label{eq: color rendering}
\end{equation}

\noindent\textbf{Video Difffusion Models.}
Diffusion models have emerged as the dominant paradigm for video generation, including two primary components. 
The forward process gradually transform a clean data sample $x_{0}\sim p(x)$ to a Gaussian noise $x_{T}\sim\mathcal{N}(0, I)$ by
\begin{equation}
    x_{t} = \sqrt{\bar{\alpha}_{t}}x_{0}+\sqrt{1-\bar{\alpha}_{t}}\epsilon, \epsilon\sim\mathcal{N}(0, I)
\end{equation}
where $x_{t}$ and $\bar{\alpha}_{t}$ denotes the noisy data and noise strength at the timestep $t$.
The reverse process removes noise from the clean data by predicting the noises:
\begin{equation}
    x_{t-1} = \dfrac{1}{\sqrt{\bar{\alpha}_{t}}}\left(x_{t}-\dfrac{\beta_{t}}{\sqrt{1-\bar{\alpha}_{t}}}\epsilon_{\theta}\left(x_{t}, t, c\right)\right)+\sigma_{t}\epsilon_{t}, \label{eq:denoise}
\end{equation}
where $c$ represents the embeddings of conditions, $\epsilon_{\theta}$ is the denoising network and $\epsilon_{t}\sim\mathcal{N}(0,I)$.
The optimization objective is defined as
\begin{equation}
    \mathcal{L}=\mathbb{E}_{\epsilon\sim\mathcal{N}(0, I), t}\left[\Vert \epsilon - \epsilon_{\theta}(x_{t},t)\Vert_{2}^{2}\right],
\end{equation}
For video generation, latent diffusion models~\cite{LDM2022CVPR} are commonly employed to mitigate the computational cost.

\begin{figure*}
    \centering
    \begin{subfigure}{0.19\linewidth}
        \centering
        \includegraphics[width=\linewidth]{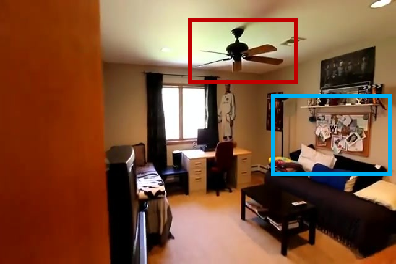}
        \caption{Reference Image}
    \end{subfigure}
    \centering
    \begin{subfigure}{0.19\linewidth}
        \centering
        \includegraphics[width=\linewidth]{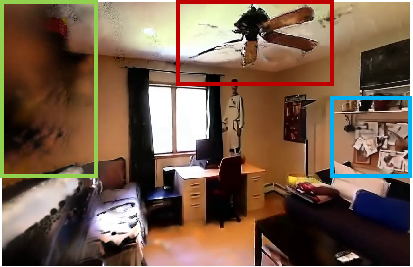}
        \caption{Flash3D~\cite{Flash3D2024NIPS}}
    \end{subfigure}
    \centering
    \begin{subfigure}{0.19\linewidth}
        \centering
        \includegraphics[width=\linewidth]{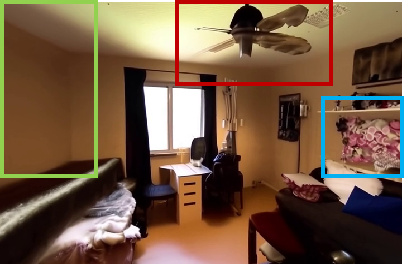}
        \caption{Video Diffusion~\cite{ViewCrafter2024arXiv}}
        \label{subfig:video diffusion}
    \end{subfigure}
    \centering
    \begin{subfigure}{0.19\linewidth}
        \centering
        \includegraphics[width=\linewidth]{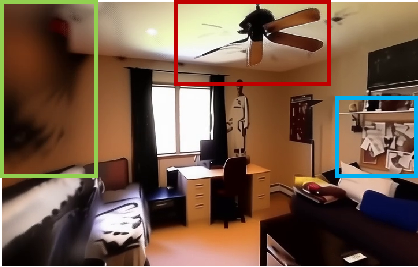}
        \caption{Latent-level Momentum}
        \label{subfig:latent-level momentum}
    \end{subfigure}
    \centering
    \begin{subfigure}{0.19\linewidth}
        \centering
        \includegraphics[width=\linewidth]{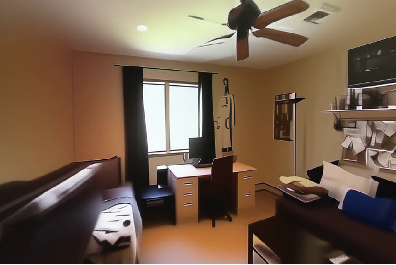}
        \caption{Two-level Momentum}
        \label{subfig:ours}
    \end{subfigure}
    \caption{Visualization of vanilla video diffusion model~\cite{ViewCrafter2024arXiv} and our two level momentum. We observe that latent-level momentum can enhance details (\textcolor{red}{red} box) while maintaining consistency (\textcolor{blue}{blue} box). Yet, such momentum limits the generation ability for unseen regions (\textcolor{green}{green} box). Motivated by this observation, we further propose pixel-level momentum to benefit from both (c) and (d).}
    \label{fig:method visual}
\end{figure*}

\subsection{Momentum Scene Generation}\label{sec:momentum scene generation}

Given the global Gaussian representations $\mathcal{G}$ and the target camera trajectory $\mathcal{K}=\{K\}_{i=1}^{N}$, we can render a video:
\begin{equation}
    \mathcal{I}=\{I_{i}\}_{i=1}^{N}=\psi_{c}(\mathcal{G}, \mathcal{K}),
\end{equation}
where $I_{i}$ is the rendering image corresponding to viewpoint $K_{i}$.
To enhance the quality of novel view rendering, we leverage latent video diffusion models (VDM)~\cite{SVD2023arXiv, CogVideoX2024arXiv, Dynamicrafter2024ECCV} trained on large-scale video datasets.
VDMs consist of a pair of VAE encoder $\mathcal{E}$ and decoder $\mathcal{D}$ to transfer videos between the RGB space and the latent space, and a video denoising network to remove noises on noisy latents.

\noindent\textbf{Latent-level Momentum.}
We encode $\mathcal{I}$ and the input image $I_{0}$ to latent space:
\begin{equation}
    \mathcal{Z}=\{Z_{i}\}_{i=1}^{N}=\mathcal{E}(\mathcal{I}), \quad Z_{0}=\mathcal{E}(I_{0}), \label{eq:encode}
\end{equation}
where $Z_{i}\in\mathbb{R}^{hw\times C}$ is the latent feature of $I_{i}$.
We introduce noisy samples direclty from $\mathcal{Z}$ as our latent-level momentum.
Then, the reverse process in Eq.~\ref{eq:denoise} can be reformulated as:
\begin{align}
    & \mathcal{Z}_{t-1}   = \lambda\left(\sqrt{\bar{\alpha}_{t-1}}\mathcal{Z}+\left(1-\sqrt{\bar{\alpha}_{t-1}}\right)\epsilon\right) \label{eq:latent momentum}
    \\ 
    &+ (1-\lambda)\left(\dfrac{1}{\sqrt{\bar{\alpha}_{t}}}\left(\mathcal{Z}_{t}-\dfrac{\beta_{t}}{\sqrt{1-\bar{\alpha}_{t}}}\epsilon_{\theta}\left(\mathcal{Z}_{t}, t, c\right)\right)+\sigma_{t}\epsilon_{t}\right),
    \nonumber
\end{align}
where $\epsilon\sim\mathcal{N}(0,I)$ is a random noise and $\lambda$ is the momentum coefficient.
Instead of treating $\lambda$ as a hyperparemeter~\cite{Repaint2022CVPR}, we design the following strategy to compute the momentum coefficient $\lambda=\{\lambda_{i}^{j}\}_{1\leq i\leq N}^{1\leq j\leq hw}\in\mathbb{R}^{N\times hw}$.
We hypothesize the first $n$ frames of $\mathcal{I}$ are well-generated, and construct a latent pool from Eq.~\ref{eq:encode}:
\begin{equation}
    \mathcal{Z}_{ref}=\{Z_{0}\}\cup \{Z_{i}\}_{i=1}^{n}\in\mathbb{R}^{(n+1)hw\times C}.
\end{equation}
We use these well-generated features as reference to compute the momentum coefficient $\lambda_{i}^{j}$ for each latent feature $z_{i}^{j}\in Z_{i}$ by:
\begin{align}
    \lambda_{i}^{j} = \max_{z\in\mathcal{Z}_{ref}}\lambda_{0}\dfrac{z\cdot z_{i}^{j}}{\Vert z\Vert\Vert z_{i}^{j}\Vert}, \label{eq:latent momentum coefficient}
\end{align}
where $\lambda_{0}\geq 0$ is an adjustable coefficient to control the overall weight of momentum.
Finally, we decode the denoised latent features $\mathcal{Z}_{0}$.
We donate the overall diffusion process as $\Phi_{\lambda}$:
\begin{equation}
    \hat{\mathcal{I}}=\Phi_{\lambda}(\mathcal{I}), \label{eq:latent momentum diffusion}
\end{equation}
where $\lambda$ is the latent-level momentum coefficient in Eq.~\ref{eq:latent momentum}.
The latent-level momentum is illustrated in Algorithm~\ref{alg: latent-level momentum}.

As shown in Figure~\ref{fig:method visual}, our latent-level momentum enables video diffusion models to generate a video with more details (in \textcolor{red}{red} boxes) and scene consistency (in \textcolor{blue}{blue} boxes).
However, such latent-level momentum limits the generation ability in unseen regions (in \textcolor{green}{green} boxes), where vanilla diffusion models can recover unseen regions better.
Therefore, we further introduce pixel-level momentum for further enhancement of unknown areas.

\begin{algorithm}[t]
\caption{latent-level Momentum}\label{alg: latent-level momentum}
\renewcommand{\algorithmicrequire}{\textbf{Input:}}
\renewcommand{\algorithmicensure}{\textbf{Output:}}
\begin{algorithmic}
\Require The reference image $I_{0}$, camera parameters $\mathcal{K}=\{K_{i}\}_{i=1}^{N}$ and Gaussian representations $\mathcal{G}$
\Ensure Enhanced frames $\tilde{\mathcal{I}}$
\State $\mathcal{I}=\{I_{i}\}_{i=1}^{N}=\psi_{c}(\mathcal{G}, \mathcal{K})$
\State $\mathcal{Z}=\{Z_{i}\}_{i=1}^{N}=\mathcal{E}(\mathcal{I}), \quad Z_{0}=\mathcal{E}(I_{0})$
\State $\lambda_{i}^{j} = \max_{z\in\mathcal{Z}_{ref}}\dfrac{z\cdot z_{i}^{j}}{\Vert z\Vert\Vert z_{i}^{j}\Vert}, \quad 1\leq i\leq N, 1\leq j\leq hw$
\For{$t\leftarrow T$ to $0$}
    \State $\epsilon\sim\mathcal{N}(0,I)$ if $t>1$ else $0$
    \State $\epsilon_{t}\sim\mathcal{N}(0,I)$ if $t>1$ else $0$
    \State $\mathcal{Z}_{t-1} = \lambda\left(\sqrt{\bar{\alpha}_{t-1}}\mathcal{Z}+\left(1-\sqrt{\bar{\alpha}_{t-1}}\right)\epsilon\right) + $
    \State $(1-\lambda)\left(\dfrac{1}{\sqrt{\bar{\alpha}_{t}}}\left(\mathcal{Z}_{t}-\dfrac{\beta_{t}}{\sqrt{1-\bar{\alpha}_{t}}}\epsilon_{\theta}\left(\mathcal{Z}_{t}, t, c\right)\right)+\sigma_{t}\epsilon_{t}\right)$
\EndFor
\State $\tilde{\mathcal{I}}=\mathcal{D}(\mathcal{Z}_{0})$
\end{algorithmic}
\end{algorithm}

\noindent\textbf{Pixel-level Momentum.}
Previous methods that directly generate videos from input views can be considered as a degenerated version ($\lambda=0$ in Eq.~\ref{eq:latent momentum diffusion}) of our latent-level momentum:
\begin{equation}
    \mathcal{I}_{new} = \Phi_{0}(\mathcal{I}),
\end{equation}
which leads to scene inconsistency (in \textcolor{blue}{blue} boxes).
Therefore, we introduce $\Phi_{\lambda}(\mathcal{I})$ from Eq.~\ref{eq:latent momentum diffusion} as pixel-level momentum to $\Phi_{0}(\mathcal{I})$:
\begin{equation}
    \mathcal{I}_{new} = \mu\Phi_{\lambda}(\mathcal{I}) + (1-\mu)\Phi_{0}(\mathcal{I}),
\end{equation}
where $\lambda$ is defined as Eq.~\ref{eq:latent momentum coefficient}.
Motivated by the observation that well-reconstructed areas are typically represented by Gaussians with very small volumes~\cite{3DGS-Enhancer2024NeurIPS}, we follow the instruction in Eq.~\ref{eq: color rendering} to render scale maps $\mathcal{S}=\{s_{i}^{j,k}\}\in[0,1)^{N\times HW\times 3}$ by
\begin{equation}
    s_{i}^{j,k}=\sum_{n\in N} (1-S_{n})\alpha_{n}\prod_{m=1}^{n-1}(1-\alpha_{n}), \label{eq:scale rendering}
\end{equation}
where $1\leq i\leq N$, $1\leq j \leq HW$ and $1\leq k\leq 3$.
Then, the pixel-level momentum coefficient $\mu=\{\mu_{i}^{j}\}\in[0,1)^{N\times HW}$ is defined as:
\begin{align}
    \mu_{i}^{j}=\left\{
    \begin{aligned}
    &\max_{k}s_{i}^{j,k}, & \tau\leq \max_{k}s_{i}^{j,k} \\
    &0, & \tau> \max_{k}s_{i}^{j,k}
    \end{aligned}
    \right.
\end{align}
where $\tau$ is a pre-defined threshold.
Higher $\mu_{i}^{j}$ represents well-reconstructed regions, which benefits more from $\Phi_{\lambda}(\mathcal{I})$, while lower $\mu_{i}^{j}$ indicates unseen regions with less details, which are further enhanced by $\Phi_{0}(\mathcal{I})$.
As shown in Figure~\ref{subfig:ours}, our method can generate high-fidelity and consistent frames with cascaded momentum.






\begin{algorithm}[t]
\caption{Overall Architecture}\label{alg: overall architecture}
\renewcommand{\algorithmicrequire}{\textbf{Input:}}
\renewcommand{\algorithmicensure}{\textbf{Output:}}
\begin{algorithmic}
\Require Input image $I_{0}$, camera parameters $\mathcal{K}=\{K_{i}\}_{i=1}^{M}$

\Ensure Gaussian parameters $\mathcal{G}=(\mu, \Sigma, \alpha, c)$
\State $\mathcal{G}=\textrm{Flash3D}(I_{0})$

\For{$s\leftarrow 0$ to $h$}
    \State $\mathcal{K}^{s}=\{K_{i}\}, \quad s(N-n)+1\leq i\leq(N-n)+N$
    \State $\mathcal{I}^{s}=\psi_{c}(\mathcal{G}, \mathcal{K}^{s})$
    \State $\mathcal{I}_{new}^{s} = \{I_{i}^{new}\}= \mu\Phi_{\lambda}(\mathcal{I}^{s}) + (1-\mu)\Phi_{0}(\mathcal{I}^{s})$
    \If{$s=0$} 
        \State $\mathcal{I}=\{I_{0}\}\cup \mathcal{I}^{0}_{new}$
    \Else
        \State $\mathcal{I}\leftarrow \mathcal{I}-\{I_{i}\}_{i=s(N-n)+1}^{s(N-n)+n}$
        \State $\mathcal{I}\leftarrow\mathcal{I}\cup \mathcal{I}^{s}_{new}$
    \EndIf
    \State $\mathcal{G} \leftarrow f_{R}(\mathcal{G}, \mathcal{I})$
\EndFor
\end{algorithmic}
\end{algorithm}

\subsection{Overall Architecture}\label{sec:overall architecture}

Given a single image $I_{0}$ and a sequence of camera parameters $\mathcal{K}=\{K_{i}\}_{i=1}^{M}$, our goal is to recover the underlying scene.
We first estimate the metric depth with a pre-trained network~\cite{UniDepth2024CVPR} and follow the instruction of Flash3D~\cite{Flash3D2024NIPS} by employing a encoder-decoder network to predict Gaussian parameters for each pixel.
In this manner, we obtain a initial representations of the scene:
\begin{equation}
    \mathcal{G}=\{(\mu_{i}, \Sigma_{i}, \alpha_{i}, c_{i})\}.
\end{equation}
Following the rendering strategy in Eq.~\ref{eq: color rendering}, we can render a video from $\mathcal{G}$ with camera parameters $\mathcal{K}^{0}=\{K_{i}\}_{i=1}^{N}$:
\begin{equation}
    \mathcal{I}^{0}=\{I_{i}\}_{i=1}^{N}=\psi^{c}(\mathcal{G}, \mathcal{K}^{0}),
\end{equation}
where $I_{i}$ is the rendering image of the unseen viewpoint $K_{i}$.
Since such rendering video $\mathcal{I}^{0}$ suffers from ambiguous geometric cues and unseen regions, we introduce a scene consistent video generation process to enhance the rendering video:
\begin{equation}
    \mathcal{I}_{new}^{0} = \mu\Phi_{\lambda}(\mathcal{I}^{0}) + (1-\mu)\Phi_{0}(\mathcal{I}^{0}).
\end{equation}
Then, we refine the scene representations based on the input image and generated video frames $\mathcal{I}=\{I_{0}\}\cup \mathcal{I}^{0}$
\begin{equation}
    \mathcal{G} \leftarrow f_{R}(\mathcal{G}, \mathcal{I}),
\end{equation}
The target is to minimize the following 3DGS loss:
\begin{equation}
    \mathcal{L}=\dfrac{1}{|\mathcal{I}|}\sum_{I\in\mathcal{I}}(1-\gamma)\mathcal{L}_{1}(\tilde{I}, I)+\gamma\mathcal{L}_{SSIM}(\tilde{I}, I),  ~\label{eq:3DGS loss}
\end{equation}
where $\mathcal{L}_{1}$, $\mathcal{L}_{SSIM}$ denote the $L_{1}$ and SSIM loss respectively, and $\gamma$ is a coefficient parameter.
In this manner, we obtain enhanced Gaussian representations for the next iteration with a new sequence of camera parameters $\mathcal{K}^{1}=\{K_{i}\}_{i=n}^{N+n}$, where $n$ is the number of overlapped frames between two steps.
Algorithm~\ref{alg: overall architecture} describes the overall architecture of our iterative reconstruction strategy.

\section{Experiments}
\begin{figure*}
    \centering
    \includegraphics[width=\linewidth]{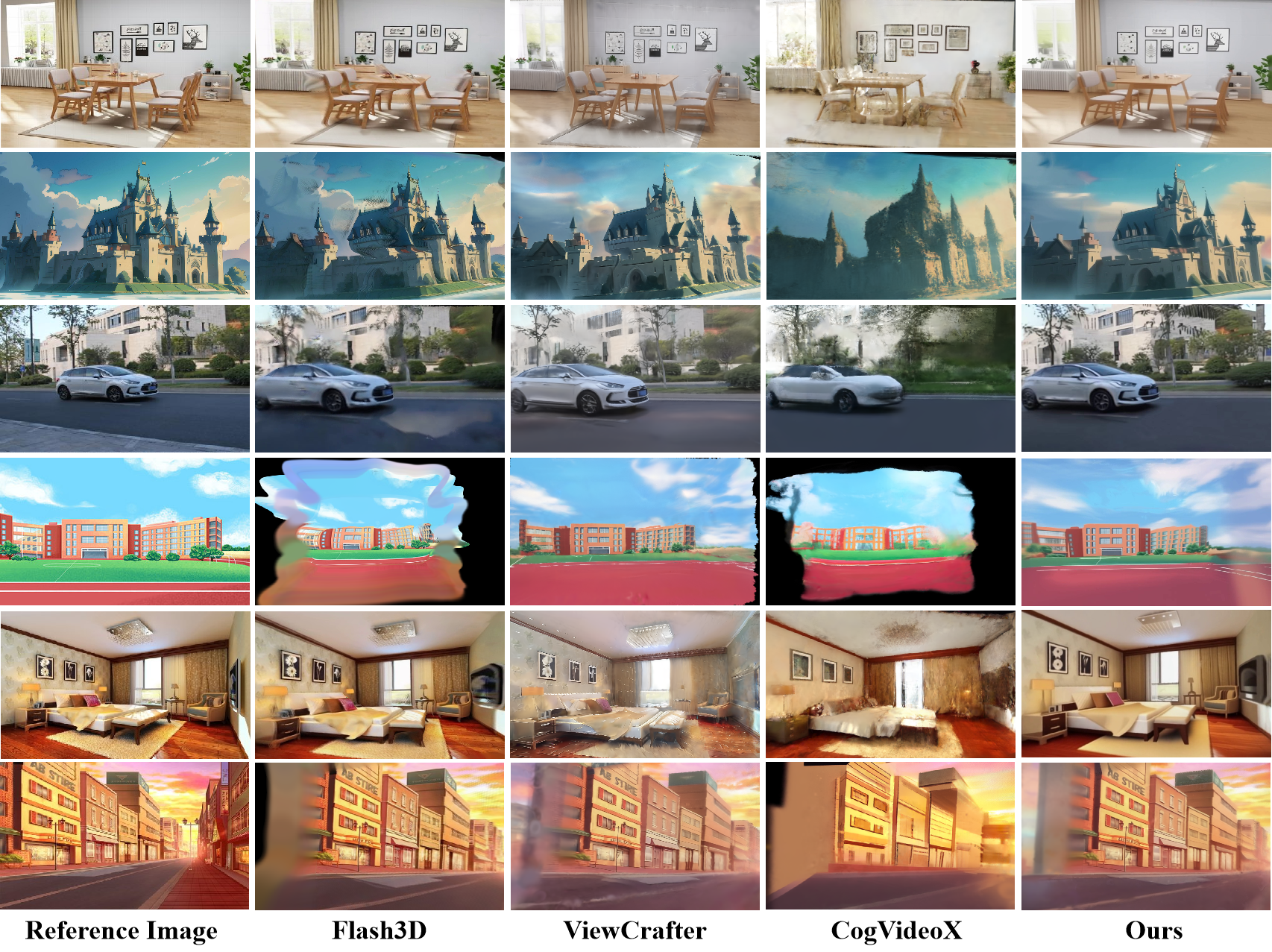}
    \caption{Qualitative comparison in 3D scene generation from single view. Given various kinds of input views, our method produces high fidelity and consistent 3D scenes.}
    \label{fig:main results}
    \vspace{-0.3cm}
\end{figure*}

We conduct extensive experiments to evaluate our method for 3D scene generation from single input.
We first present the setup of our experiments in Section~\ref{sec: experiment setup}.
Then, we report our qualitative and quantitative results compared to representative baseline methods in Section~\ref{sec:3D scene generation}.
We further analyze the iterative scene reconstruction process in Section~\ref{sec:iterative scene reconstruction}.
Finally, we conduct ablation studies to verify the efficacy of our framework design in Section~\ref{sec:ablation}.

\subsection{Experimental setup}\label{sec: experiment setup}
In our framework, we first adopt Flash3D~\cite{Flash3D2024NIPS} to predict Gaussian representations from the input image as initialization. 
The video diffusion model used for $\Phi_{\lambda}$ and $\Phi_{0}$ is trained by ViewCrafter~\cite{ViewCrafter2024arXiv}, with a CLIP image encoder for input image understanding.
The length of videos in each step is set to $N=25$ and the number of overlapped frames between two steps is set to $n=10$.
The number of total iterations $h$ is adapted to the length of camera trajectory.
In each iteration, we optimize Gaussian representations for $5000$ steps with $\gamma=0.2$.
The densification interval is set to $100$, and the opacity of Gaussians is reset every $3000$ steps. 

For 3D scene generation from single image, we compare our methods with one regression-based method~\cite{Flash3D2024NIPS} and two generative-based methods~\cite{CogVideoX2024arXiv, ViewCrafter2024arXiv}.
Specifically, Flash3D~\cite{Flash3D2024NIPS} uses a frozen off-the-shelf network~\cite{UniDepth2024CVPR} for depth estimation to predict the first layer of Gaussians and adds addtional layers of Gaussians that are offset in space.
The model is pre-trained on the RealEstate10k~\cite{Re10k2018arXiv} dataset.
CogVideoX~\cite{CogVideoX2024arXiv} is a large-scale video generation model based on diffusion transformer, trained on 35M video clips and 2B images.
We adopt the most powerful version of CogVideoX~\cite{CogVideoX2024arXiv}, which has 5B parameters.
ViewCrafter~\cite{ViewCrafter2024arXiv} introduces a point-conditioned video diffusion and iterative camera trajectory.
For quantitative results, We construct a subset of RealEstate10K~\cite{Re10k2018arXiv} for evaluation.
The easy test set has smaller movement of viewpoints, and the hard set has larger view ranges.
We employ PSNR, SSIM~\cite{SSIM2004TIP} and LPIPS~\cite{LPIPS2018CVPR} as the evaluation metrics for rendering image quality.

\begin{table*}[t]
    \centering
    \caption{Quantitative comparison in 3D scene generation from single view. We report the average results of PSNR, SSIM and LPIPS. Our approach ourperforms other baselines in all metrix.}
    \setlength{\tabcolsep}{4.1mm}{\begin{tabular}{c|cccccc}
        \toprule
        \multirow{2}{*}{Method}  & \multicolumn{3}{c}{Easy Set} & \multicolumn{3}{c}{Hard Set} \\
        \cmidrule(lr){2-4} \cmidrule(lr){5-7}
         & PSNR↑ & SSIM↑ & LPIPS↓   & PSNR↑ & SSIM↑ & LPIPS↓ \\
        \cmidrule(lr){1-7}
        Flash3D~\cite{Flash3D2024NIPS} & 17.94 & 0.682 & 0.160 & 14.41 & 0.599 & 0.370 \\
        CogVideoX~\cite{CogVideoX2024arXiv} & 17.25 & 0.710 & 0.352& 15.42 & 0.636 & 0.415 \\
        ViewCrafter~\cite{ViewCrafter2024arXiv} & 20.70 & 0.794 & 0.159 & 15.63 & 0.676 & 0.258 \\
        Ours & \textbf{20.95} & \textbf{0.800} & \textbf{0.145} & \textbf{17.62} & \textbf{0.707} & \textbf{0.233} \\
        \bottomrule
    \end{tabular}}
    \label{tab:quantitative comparison}
\end{table*}

\subsection{3D Scene Generation}~\label{sec:3D scene generation}
\noindent\textbf{Qualitative Results.}
As shown in Figure~\ref{fig:main results}, we visualize the final rendering results of our method and baselines.
Unambiguous geometric cues are unavailable for Flash3D~\cite{Flash3D2024NIPS} from single input, leading to distortions of the table and chairs in row 1, column 2.
Besides, it also lack the ability to recover unseen regions, which is obvious in the zoom-out setting, as shown in row 4, column 2.
ViewCrafter~\cite{ViewCrafter2024arXiv} and CogVideoX~\cite{CogVideoX2024arXiv} can enhance the input frames, but suffer from scene inconsistency, which leads to conflicts in further reconstruction.
For example, CogVideoX~\cite{CogVideoX2024arXiv} generates different chairs compared to the input image in row 1, and ViewCrafter~\cite{ViewCrafter2024arXiv} changes the color style of the scene in row 3.
Our cascaded momentum can provide high quality observations while maintaining the scene consistency.
The various styles of inputs, from cartoon to realistic images, from indoor to outdoor scenes, also demonstrate the generalization ability of our method.

\noindent\textbf{Quantitative Results.}
The quantitative comparison results are reported in Table~\ref{tab:quantitative comparison}. 
For the easy test set, the simple employment of CogVideoX~\cite{CogVideoX2024arXiv} leads to worse performance in PSNR and LPIPS, which demonstrates the negative effects of inconsistent generation.
With the same video diffusion model, our method outperforms ViewCrafter~\cite{ViewCrafter2024arXiv} in all matrix on the easy set, and shows greater advantages on the hard set.
The lower LPIPS score further demonstrates that our approach generates more perceptually accurate images.

\begin{figure*}
    \centering
    \includegraphics[width=\linewidth]{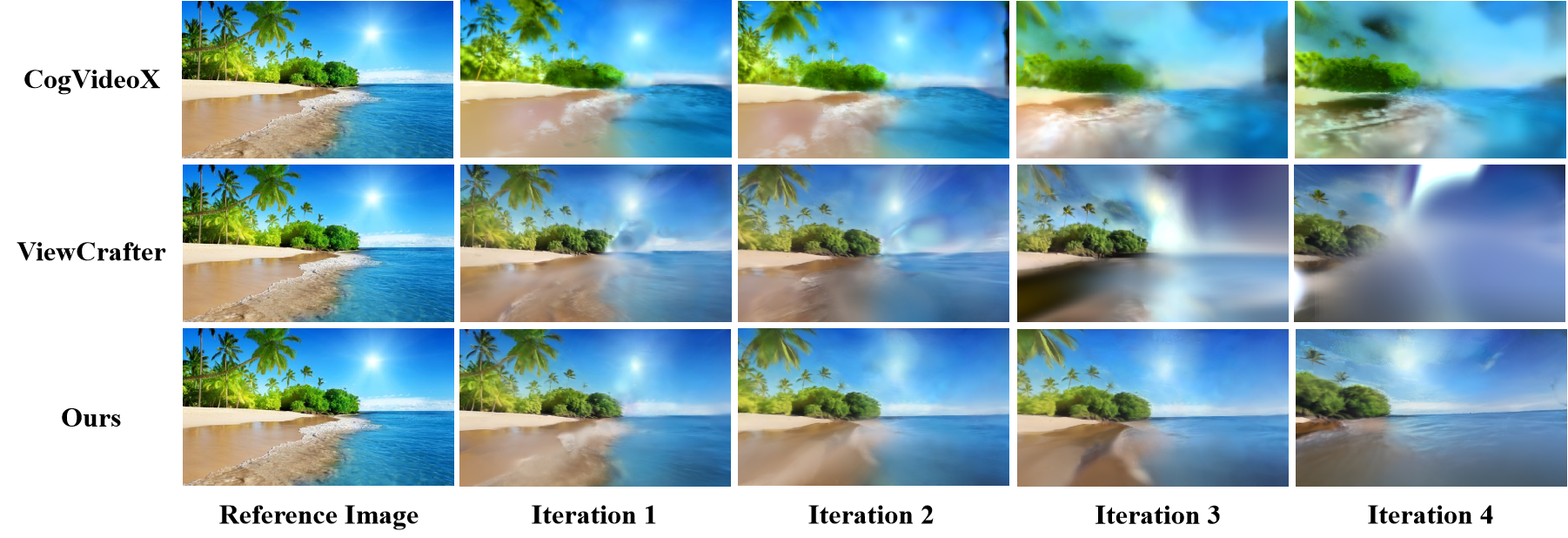}
    \caption{Visualization of rendering results in each iteration. The inconsistency in CogVideoX~\cite{CogVideoX2024arXiv} and ViewCrafter~\cite{ViewCrafter2024arXiv} gradually increases. Our method can maintain high consistency during the iterative reconstruction process.}
    \label{fig:iter recon}
    \vspace{-0.5cm}
\end{figure*}

\begin{figure}
    \centering
    \includegraphics[width=\linewidth]{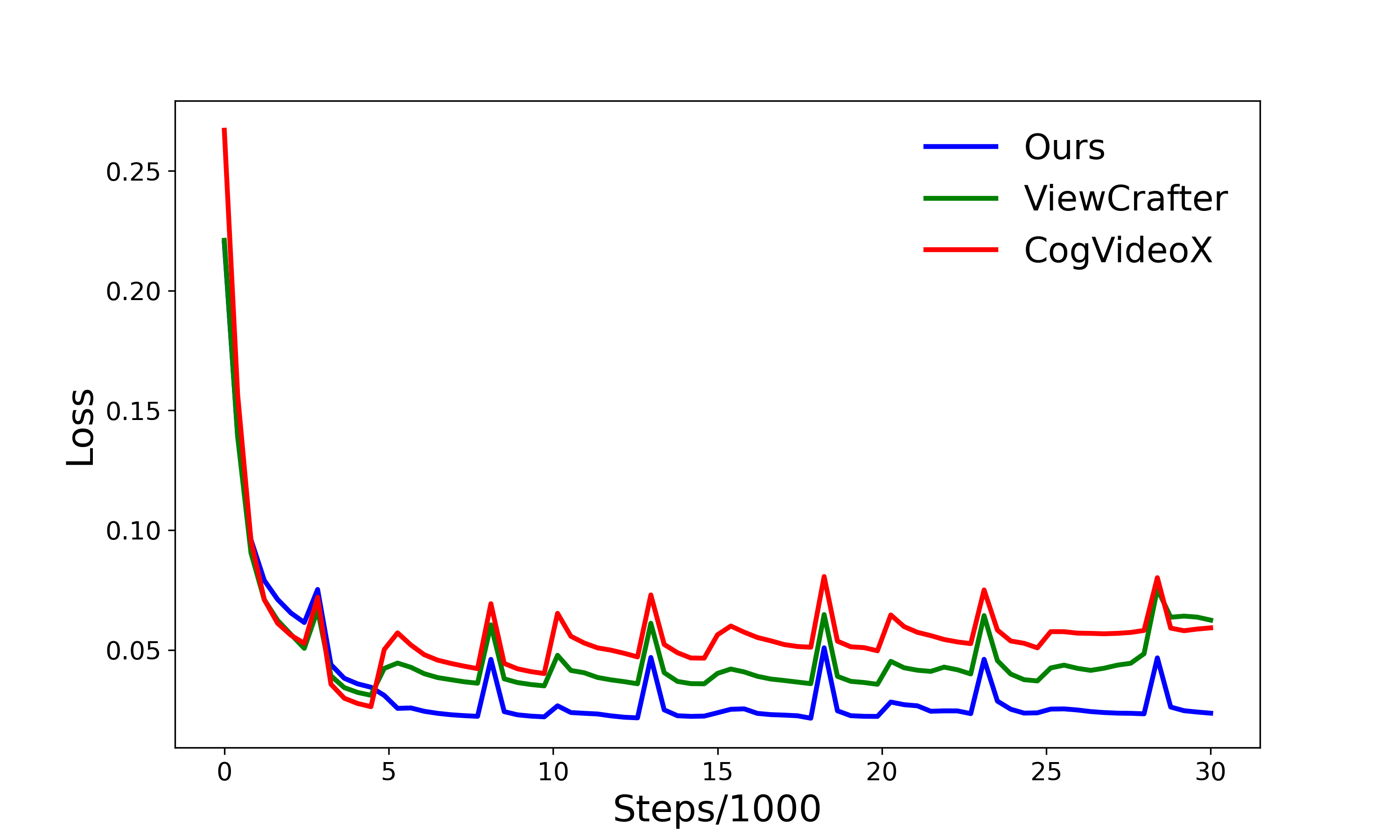}
    \caption{Reconstruction loss in optimization of Gaussian representations. The overall process includes 5 iterations, where we optimize 3DGS for 5000 steps in each iteration.}
    \vspace{-0.5cm}
    \label{fig:recon loss}

\end{figure}

\subsection{Iterative Gaussian Reconstruction}~\label{sec:iterative scene reconstruction}
\noindent Our paradigm follows an iterative Gaussian reconstruction process to avoid the restriction of video length in video diffusion models.
Under this circumstance, the inconsistency of generated videos will be accumulated in each iteration.
As shown in Figure~\ref{fig:iter recon}, although ViewCrafter~\cite{ViewCrafter2024arXiv} and CogVideoX~\cite{CogVideoX2024arXiv} maintain decent performance in the first two iterations, they fail to reconstruct a reasonable scene in further iterations with larger view ranges.
Benefiting from cascaded momentum, our method maintains scene consistency during iterative reconstruction for the global Gaussian representations.
We also visualize the 3DGS loss in Figure~\ref{fig:recon loss}.
The reconstruction process in CogVideoX~\cite{CogVideoX2024arXiv} and ViewCrafter~\cite{ViewCrafter2024arXiv} can not fully converge due to the inconsistency between different frames. 
At the end of reconstruction, their losses are even higher than those at the end of the first step.
In contrast, our method can converge to a lower loss value at each iteration with more consistent novel views as supervision information of Gaussian optimization.

\begin{figure*}
    \centering
    \includegraphics[width=\linewidth]{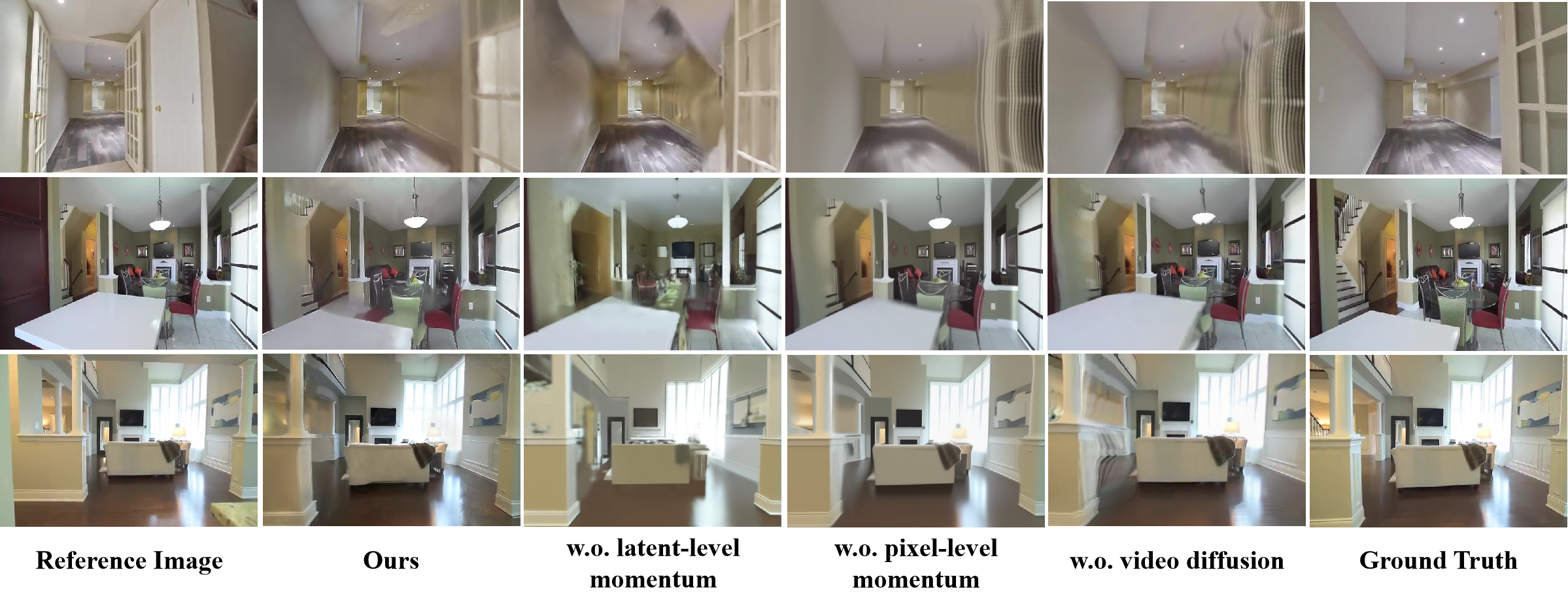}
    \caption{Visualization of the ablation study. }
    \label{fig:ablation}
    \vspace{-0.2cm}
\end{figure*}


\subsection{Ablation Study}~\label{sec:ablation}
\noindent We conduct ablation studies to investigate our momentum-based 3D generation paradigm. 
We report the quantitative results in Table~\ref{tab:ablation} and visualize the rendering results in Figure~\ref{fig:ablation}.
Without any generative prior knowledge, our framework degenerates to Flash3D~\cite{Flash3D2024NIPS}, where rendering results in column 5 suffer from distortions in geometry, since the estimation of depth is not fully supervised in monocular settings.
This issue leads to a decrease of 3.67dB in PSNR and 0.126 in SSIM.
The third column in Figure~\ref{fig:ablation} illustrates that the absence of latent-level momentum results in the change of existing components.
Such experimental results demonstrate the ability of latent-level momentum to preserve scene consistency.
We also remove the pixel-level momentum in our paradigm. 
Lack of pixel-level momentum restricts the generation ability of diffusion models, with a decrease of 3.5dB in PSNR and 0.111 in SSIM.

\section{Conclusion and Discussion}

In this paper, we propose a momentum-based paradigm to generate 3D scene from single image.
We generate noisy samples from the original features to serve as momentum, thereby enhancing video details and preserving scene consistency. Nevertheless, when it comes to latent features with a perception field that spans both known and unknown regions, the latent-level momentum can impede the generative capacity within the unknown areas.
To overcome this limitation, we incorporate the previously mentioned consistent video as a pixel-level momentum to a video generated directly without momentum, aiming to achieve a more effective recovery of unseen regions. 
This cascaded momentum strategy empowers video diffusion models to produce both high-fidelity and consistent novel views.
We further refine the global Gaussian representations using the enhanced frames and render new frames for the momentum update in the subsequent step. 
This iterative process allows us to reconstruct a 3D scene, bypassing the constraints typically associated with video length.
Comprehensive experiments have demonstrated the generalization capability and the superior performance of our method in generating scenes that are both high-fidelity and consistent.

\begin{table}[h]
    \centering
    \caption{Ablation study of our Scene Splatter. We report the average PSNR, SSIM and LPIPS of rendering results.}
    \begin{tabular}{l|ccc}
    \toprule
    Methods & PSNR & SSIM & LPIPS \\
    \midrule
    Ours    & 19.81 & 0.770 & 0.169\\
    w.o. video diffusion model & 16.14 & 0.644 & 0.268 \\
    w.o. latent-level momentum & 18.52 & 0.748 & 0.194 \\
    w.o. pixel-level momentum & 16.31 & 0.659 & 0.278 \\
    \bottomrule
    \end{tabular}
    \label{tab:ablation}
    \vspace{-0.3cm}
\end{table}

\noindent\textbf{Limitations and Future Works.}
Although our method can generate high fidelity and consistent 3D scene from one single image, the employment of video diffusion models requires longer time compared to regression-based models (\textit{e.g.} Flash3D~\cite{Flash3D2024NIPS}). The iterative strategy further exacerbates the consumption of time.
Besides, our method is now restricted to static 3D scene generation, where our 3D representations lack the ability to recover 4D scenes.
In the future, we will focus on the efficiency of 3D scene generation models to further compress the time consumption. We are also interested in generating 4D scenes by decoupling the temporal and spatial factors in video diffusion.

\noindent\textbf{Acknowledgments}
This work was supported in part by the National Natural Science Foundation of China under Grant 62206147, and in part by 2024 WeChat Vision, Tecent Inc. Rhino-Bird Focused Research Program.




\begin{thebibliography}{48}
\providecommand{\natexlab}[1]{#1}
\providecommand{\url}[1]{\texttt{#1}}
\expandafter\ifx\csname urlstyle\endcsname\relax
  \providecommand{\doi}[1]{doi: #1}\else
  \providecommand{\doi}{doi: \begingroup \urlstyle{rm}\Url}\fi

\bibitem[Blattmann et~al.(2023)Blattmann, Dockhorn, Kulal, Mendelevitch, Kilian, Lorenz, Levi, English, Voleti, Letts, et~al.]{SVD2023arXiv}
Andreas Blattmann, Tim Dockhorn, Sumith Kulal, Daniel Mendelevitch, Maciej Kilian, Dominik Lorenz, Yam Levi, Zion English, Vikram Voleti, Adam Letts, et~al.
\newblock Stable video diffusion: Scaling latent video diffusion models to large datasets.
\newblock \emph{arXiv preprint arXiv:2311.15127}, 2023.

\bibitem[Charatan et~al.(2024)Charatan, Li, Tagliasacchi, and Sitzmann]{pixelSplat2024CVPR}
David Charatan, Sizhe~Lester Li, Andrea Tagliasacchi, and Vincent Sitzmann.
\newblock pixelsplat: 3d gaussian splats from image pairs for scalable generalizable 3d reconstruction.
\newblock In \emph{CVPR}, pages 19457--19467, 2024.

\bibitem[Chen et~al.(2021)Chen, Xu, Zhao, Zhang, Xiang, Yu, and Su]{MVSNeRF2021ICCV}
Anpei Chen, Zexiang Xu, Fuqiang Zhao, Xiaoshuai Zhang, Fanbo Xiang, Jingyi Yu, and Hao Su.
\newblock Mvsnerf: Fast generalizable radiance field reconstruction from multi-view stereo.
\newblock In \emph{ICCV}, pages 14124--14133, 2021.

\bibitem[Chen et~al.(2024{\natexlab{a}})Chen, Zhang, Cun, Xia, Wang, Weng, and Shan]{Videocrafter2024arXiv}
Haoxin Chen, Yong Zhang, Xiaodong Cun, Menghan Xia, Xintao Wang, Chao Weng, and Ying Shan.
\newblock Videocrafter2: Overcoming data limitations for high-quality video diffusion models, 2024{\natexlab{a}}.

\bibitem[Chen et~al.(2024{\natexlab{b}})Chen, Xu, Zheng, Zhuang, Pollefeys, Geiger, Cham, and Cai]{MVSplat2024ECCV}
Yuedong Chen, Haofei Xu, Chuanxia Zheng, Bohan Zhuang, Marc Pollefeys, Andreas Geiger, Tat-Jen Cham, and Jianfei Cai.
\newblock Mvsplat: Efficient 3d gaussian splatting from sparse multi-view images.
\newblock In \emph{ECCV}, pages 370--386. Springer, 2024{\natexlab{b}}.

\bibitem[Chen et~al.(2024{\natexlab{c}})Chen, Wang, Wang, Wang, and Liu]{V3D2024arXiv}
Zilong Chen, Yikai Wang, Feng Wang, Zhengyi Wang, and Huaping Liu.
\newblock V3d: Video diffusion models are effective 3d generators.
\newblock \emph{arXiv preprint arXiv:2403.06738}, 2024{\natexlab{c}}.

\bibitem[Chibane et~al.(2021)Chibane, Bansal, Lazova, and Pons-Moll]{SRF2021CVPR}
Julian Chibane, Aayush Bansal, Verica Lazova, and Gerard Pons-Moll.
\newblock Stereo radiance fields (srf): Learning view synthesis for sparse views of novel scenes.
\newblock In \emph{CVPR}, pages 7911--7920, 2021.

\bibitem[Fridman et~al.(2023)Fridman, Abecasis, Kasten, and Dekel]{ScenScape2023}
Rafail Fridman, Amit Abecasis, Yoni Kasten, and Tali Dekel.
\newblock Scenescape: Text-driven consistent scene generation.
\newblock \emph{NeurIPS}, 36, 2023.

\bibitem[He et~al.(2024)He, Xu, Guo, Wetzstein, Dai, Li, and Yang]{CameraCtrl2024arXiv}
Hao He, Yinghao Xu, Yuwei Guo, Gordon Wetzstein, Bo Dai, Hongsheng Li, and Ceyuan Yang.
\newblock Cameractrl: Enabling camera control for text-to-video generation.
\newblock \emph{arXiv preprint arXiv:2404.02101}, 2024.

\bibitem[He et~al.(2022)He, Yang, Zhang, Shan, and Chen]{LVDM2022arXiv}
Yingqing He, Tianyu Yang, Yong Zhang, Ying Shan, and Qifeng Chen.
\newblock Latent video diffusion models for high-fidelity long video generation.
\newblock \emph{arXiv}, 2022.

\bibitem[Ho et~al.(2020)Ho, Jain, and Abbeel]{DDPM2020NIPS}
Jonathan Ho, Ajay Jain, and Pieter Abbeel.
\newblock Denoising diffusion probabilistic models.
\newblock \emph{NeurIPS}, 33:\penalty0 6840--6851, 2020.

\bibitem[Kerbl et~al.(2023)Kerbl, Kopanas, Leimk{\"u}hler, and Drettakis]{3DGS2023ToG}
Bernhard Kerbl, Georgios Kopanas, Thomas Leimk{\"u}hler, and George Drettakis.
\newblock 3d gaussian splatting for real-time radiance field rendering.
\newblock \emph{ACM Trans. Graph.}, 42\penalty0 (4):\penalty0 139--1, 2023.

\bibitem[Lee et~al.(2024)Lee, Chen, Wang, Liao, Feng, and Huang]{VividDream2024arXiv}
Yao-Chih Lee, Yi-Ting Chen, Andrew Wang, Ting-Hsuan Liao, Brandon~Y Feng, and Jia-Bin Huang.
\newblock Vividdream: Generating 3d scene with ambient dynamics.
\newblock \emph{arXiv preprint arXiv:2405.20334}, 2024.

\bibitem[Li et~al.(2023{\natexlab{a}})Li, Tan, Zhang, Xu, Luan, Xu, Hong, Sunkavalli, Shakhnarovich, and Bi]{Instant3d2023arXiv}
Jiahao Li, Hao Tan, Kai Zhang, Zexiang Xu, Fujun Luan, Yinghao Xu, Yicong Hong, Kalyan Sunkavalli, Greg Shakhnarovich, and Sai Bi.
\newblock Instant3d: Fast text-to-3d with sparse-view generation and large reconstruction model.
\newblock \emph{arXiv preprint arXiv:2311.06214}, 2023{\natexlab{a}}.

\bibitem[Li et~al.(2023{\natexlab{b}})Li, Chen, Chen, and Tan]{Sweetdreamer2023arXiv}
Weiyu Li, Rui Chen, Xuelin Chen, and Ping Tan.
\newblock Sweetdreamer: Aligning geometric priors in 2d diffusion for consistent text-to-3d.
\newblock \emph{arXiv preprint arXiv:2310.02596}, 2023{\natexlab{b}}.

\bibitem[Lin et~al.(2023)Lin, Gao, Tang, Takikawa, Zeng, Huang, Kreis, Fidler, Liu, and Lin]{Magic3d2023CVPR}
Chen-Hsuan Lin, Jun Gao, Luming Tang, Towaki Takikawa, Xiaohui Zeng, Xun Huang, Karsten Kreis, Sanja Fidler, Ming-Yu Liu, and Tsung-Yi Lin.
\newblock Magic3d: High-resolution text-to-3d content creation.
\newblock In \emph{CVPR}, pages 300--309, 2023.

\bibitem[Liu et~al.(2023{\natexlab{a}})Liu, Xu, Jin, Chen, Varma~T, Xu, and Su]{One23452023NeurIPS}
Minghua Liu, Chao Xu, Haian Jin, Linghao Chen, Mukund Varma~T, Zexiang Xu, and Hao Su.
\newblock One-2-3-45: Any single image to 3d mesh in 45 seconds without per-shape optimization.
\newblock \emph{NeurIPS}, 36, 2023{\natexlab{a}}.

\bibitem[Liu et~al.(2023{\natexlab{b}})Liu, Wu, Van~Hoorick, Tokmakov, Zakharov, and Vondrick]{Zero1232023ICCV}
Ruoshi Liu, Rundi Wu, Basile Van~Hoorick, Pavel Tokmakov, Sergey Zakharov, and Carl Vondrick.
\newblock Zero-1-to-3: Zero-shot one image to 3d object.
\newblock In \emph{ICCV}, pages 9298--9309, 2023{\natexlab{b}}.

\bibitem[Liu et~al.(2024)Liu, Zhou, and Huang]{3DGS-Enhancer2024NeurIPS}
Xi Liu, Chaoyi Zhou, and Siyu Huang.
\newblock 3dgs-enhancer: Enhancing unbounded 3d gaussian splatting with view-consistent 2d diffusion priors.
\newblock \emph{NeurIPS}, 2024.

\bibitem[Long et~al.(2024)Long, Guo, Lin, Liu, Dou, Liu, Ma, Zhang, Habermann, Theobalt, et~al.]{Wonder3D2024CVPR}
Xiaoxiao Long, Yuan-Chen Guo, Cheng Lin, Yuan Liu, Zhiyang Dou, Lingjie Liu, Yuexin Ma, Song-Hai Zhang, Marc Habermann, Christian Theobalt, et~al.
\newblock Wonder3d: Single image to 3d using cross-domain diffusion.
\newblock In \emph{CVPR}, pages 9970--9980, 2024.

\bibitem[Lugmayr et~al.(2022)Lugmayr, Danelljan, Romero, Yu, Timofte, and Van~Gool]{Repaint2022CVPR}
Andreas Lugmayr, Martin Danelljan, Andres Romero, Fisher Yu, Radu Timofte, and Luc Van~Gool.
\newblock Repaint: Inpainting using denoising diffusion probabilistic models.
\newblock In \emph{CVPR}, pages 11461--11471, 2022.

\bibitem[Mildenhall et~al.(2021)Mildenhall, Srinivasan, Tancik, Barron, Ramamoorthi, and Ng]{NeRF2021ACM}
Ben Mildenhall, Pratul~P Srinivasan, Matthew Tancik, Jonathan~T Barron, Ravi Ramamoorthi, and Ren Ng.
\newblock Nerf: Representing scenes as neural radiance fields for view synthesis.
\newblock \emph{Communications of the ACM}, 65\penalty0 (1):\penalty0 99--106, 2021.

\bibitem[Piccinelli et~al.(2024)Piccinelli, Yang, Sakaridis, Segu, Li, Van~Gool, and Yu]{UniDepth2024CVPR}
Luigi Piccinelli, Yung-Hsu Yang, Christos Sakaridis, Mattia Segu, Siyuan Li, Luc Van~Gool, and Fisher Yu.
\newblock Unidepth: Universal monocular metric depth estimation.
\newblock In \emph{CVPR}, pages 10106--10116, 2024.

\bibitem[Ramesh et~al.(2022)Ramesh, Dhariwal, Nichol, Chu, and Chen]{HierarchicalT2V2022}
Aditya Ramesh, Prafulla Dhariwal, Alex Nichol, Casey Chu, and Mark Chen.
\newblock Hierarchical text-conditional image generation with clip latents.
\newblock \emph{arXiv preprint arXiv:2204.06125}, 1\penalty0 (2):\penalty0 3, 2022.

\bibitem[Rematas et~al.(2021)Rematas, Martin-Brualla, and Ferrari]{Sharf2021CVPR}
Konstantinos Rematas, Ricardo Martin-Brualla, and Vittorio Ferrari.
\newblock Sharf: Shape-conditioned radiance fields from a single view.
\newblock \emph{CVPR}, 2021.

\bibitem[Rombach et~al.(2022)Rombach, Blattmann, Lorenz, Esser, and Ommer]{LDM2022CVPR}
Robin Rombach, Andreas Blattmann, Dominik Lorenz, Patrick Esser, and Bj{\"o}rn Ommer.
\newblock High-resolution image synthesis with latent diffusion models.
\newblock In \emph{CVPR}, pages 10684--10695, 2022.

\bibitem[Saharia et~al.(2022)Saharia, Chan, Saxena, Li, Whang, Denton, Ghasemipour, Gontijo~Lopes, Karagol~Ayan, Salimans, et~al.]{PhotorealisticT2I2022NIPS}
Chitwan Saharia, William Chan, Saurabh Saxena, Lala Li, Jay Whang, Emily~L Denton, Kamyar Ghasemipour, Raphael Gontijo~Lopes, Burcu Karagol~Ayan, Tim Salimans, et~al.
\newblock Photorealistic text-to-image diffusion models with deep language understanding.
\newblock \emph{NeurIPS}, 35:\penalty0 36479--36494, 2022.

\bibitem[Shi et~al.(2023)Shi, Wang, Ye, Long, Li, and Yang]{MVdream2023arXiv}
Yichun Shi, Peng Wang, Jianglong Ye, Mai Long, Kejie Li, and Xiao Yang.
\newblock Mvdream: Multi-view diffusion for 3d generation.
\newblock \emph{arXiv preprint arXiv:2308.16512}, 2023.

\bibitem[Shriram et~al.(2024)Shriram, Trevithick, Liu, and Ramamoorthi]{RealmDreamer2024arXiv}
Jaidev Shriram, Alex Trevithick, Lingjie Liu, and Ravi Ramamoorthi.
\newblock Realmdreamer: Text-driven 3d scene generation with inpainting and depth diffusion.
\newblock \emph{arXiv preprint arXiv:2404.07199}, 2024.

\bibitem[Song et~al.(2021)Song, Meng, and Ermon]{DDIM2021ICLR}
Jiaming Song, Chenlin Meng, and Stefano Ermon.
\newblock Denoising diffusion implicit models.
\newblock \emph{ICLR}, 2021.

\bibitem[Szymanowicz et~al.(2024{\natexlab{a}})Szymanowicz, Insafutdinov, Zheng, Campbell, Henriques, Rupprecht, and Vedaldi]{Flash3D2024NIPS}
Stanislaw Szymanowicz, Eldar Insafutdinov, Chuanxia Zheng, Dylan Campbell, Jo{\~a}o~F Henriques, Christian Rupprecht, and Andrea Vedaldi.
\newblock Flash3d: Feed-forward generalisable 3d scene reconstruction from a single image.
\newblock \emph{NeurIPS}, 2024{\natexlab{a}}.

\bibitem[Szymanowicz et~al.(2024{\natexlab{b}})Szymanowicz, Rupprecht, and Vedaldi]{Splatterimage2024CVPR}
Stanislaw Szymanowicz, Chrisitian Rupprecht, and Andrea Vedaldi.
\newblock Splatter image: Ultra-fast single-view 3d reconstruction.
\newblock In \emph{CVPR}, pages 10208--10217, 2024{\natexlab{b}}.

\bibitem[Tang et~al.(2024)Tang, Chen, Chen, Wang, Zeng, and Liu]{LGM2024ECCV}
Jiaxiang Tang, Zhaoxi Chen, Xiaokang Chen, Tengfei Wang, Gang Zeng, and Ziwei Liu.
\newblock Lgm: Large multi-view gaussian model for high-resolution 3d content creation.
\newblock In \emph{ECCV}, pages 1--18. Springer, 2024.

\bibitem[Trevithick and Yang(2021)]{Grf2021ICCV}
Alex Trevithick and Bo Yang.
\newblock Grf: Learning a general radiance field for 3d representation and rendering.
\newblock In \emph{ICCV}, pages 15182--15192, 2021.

\bibitem[Voleti et~al.(2024)Voleti, Yao, Boss, Letts, Pankratz, Tochilkin, Laforte, Rombach, and Jampani]{SV3D2024ECCV}
Vikram Voleti, Chun-Han Yao, Mark Boss, Adam Letts, David Pankratz, Dmitry Tochilkin, Christian Laforte, Robin Rombach, and Varun Jampani.
\newblock Sv3d: Novel multi-view synthesis and 3d generation from a single image using latent video diffusion.
\newblock In \emph{ECCV}, pages 439--457. Springer, 2024.

\bibitem[Wang et~al.(2024{\natexlab{a}})Wang, Huang, Chen, and Lee]{FreeSplat2024NIPS}
Yunsong Wang, Tianxin Huang, Hanlin Chen, and Gim~Hee Lee.
\newblock Freesplat: Generalizable 3d gaussian splatting towards free-view synthesis of indoor scenes.
\newblock \emph{NeurIPS}, 2024{\natexlab{a}}.

\bibitem[Wang et~al.(2004)Wang, Bovik, Sheikh, and Simoncelli]{SSIM2004TIP}
Zhou Wang, Alan~C Bovik, Hamid~R Sheikh, and Eero~P Simoncelli.
\newblock Image quality assessment: from error visibility to structural similarity.
\newblock \emph{TIP}, 13\penalty0 (4):\penalty0 600--612, 2004.

\bibitem[Wang et~al.(2024{\natexlab{b}})Wang, Lu, Wang, Bao, Li, Su, and Zhu]{Prolificdreamer2024NIPS}
Zhengyi Wang, Cheng Lu, Yikai Wang, Fan Bao, Chongxuan Li, Hang Su, and Jun Zhu.
\newblock Prolificdreamer: High-fidelity and diverse text-to-3d generation with variational score distillation.
\newblock \emph{NeurIPS}, 36, 2024{\natexlab{b}}.

\bibitem[Wang et~al.(2024{\natexlab{c}})Wang, Yuan, Wang, Li, Chen, Xia, Luo, and Shan]{MotionCtrl2024SIGGRAPH}
Zhouxia Wang, Ziyang Yuan, Xintao Wang, Yaowei Li, Tianshui Chen, Menghan Xia, Ping Luo, and Ying Shan.
\newblock Motionctrl: A unified and flexible motion controller for video generation.
\newblock In \emph{SIGGRAPH}, pages 1--11, 2024{\natexlab{c}}.

\bibitem[Wu et~al.(2024)Wu, Mildenhall, Henzler, Park, Gao, Watson, Srinivasan, Verbin, Barron, Poole, et~al.]{Reconfusion2024CVPR}
Rundi Wu, Ben Mildenhall, Philipp Henzler, Keunhong Park, Ruiqi Gao, Daniel Watson, Pratul~P Srinivasan, Dor Verbin, Jonathan~T Barron, Ben Poole, et~al.
\newblock Reconfusion: 3d reconstruction with diffusion priors.
\newblock In \emph{CVPR}, pages 21551--21561, 2024.

\bibitem[Xing et~al.(2024)Xing, Xia, Zhang, Chen, Yu, Liu, Liu, Wang, Shan, and Wong]{Dynamicrafter2024ECCV}
Jinbo Xing, Menghan Xia, Yong Zhang, Haoxin Chen, Wangbo Yu, Hanyuan Liu, Gongye Liu, Xintao Wang, Ying Shan, and Tien-Tsin Wong.
\newblock Dynamicrafter: Animating open-domain images with video diffusion priors.
\newblock In \emph{ECCV}, pages 399--417. Springer, 2024.

\bibitem[Yang et~al.(2024)Yang, Teng, Zheng, Ding, Huang, Xu, Yang, Hong, Zhang, Feng, et~al.]{CogVideoX2024arXiv}
Zhuoyi Yang, Jiayan Teng, Wendi Zheng, Ming Ding, Shiyu Huang, Jiazheng Xu, Yuanming Yang, Wenyi Hong, Xiaohan Zhang, Guanyu Feng, et~al.
\newblock Cogvideox: Text-to-video diffusion models with an expert transformer.
\newblock \emph{arXiv preprint arXiv:2408.06072}, 2024.

\bibitem[Yu et~al.(2021)Yu, Ye, Tancik, and Kanazawa]{pixelNeRF2021CVPR}
Alex Yu, Vickie Ye, Matthew Tancik, and Angjoo Kanazawa.
\newblock pixelnerf: Neural radiance fields from one or few images.
\newblock In \emph{CVPR}, pages 4578--4587, 2021.

\bibitem[Yu et~al.(2024{\natexlab{a}})Yu, Duan, Hur, Sargent, Rubinstein, Freeman, Cole, Sun, Snavely, Wu, et~al.]{WonderJourney2024CVPR}
Hong-Xing Yu, Haoyi Duan, Junhwa Hur, Kyle Sargent, Michael Rubinstein, William~T Freeman, Forrester Cole, Deqing Sun, Noah Snavely, Jiajun Wu, et~al.
\newblock Wonderjourney: Going from anywhere to everywhere.
\newblock In \emph{CVPR}, pages 6658--6667, 2024{\natexlab{a}}.

\bibitem[Yu et~al.(2024{\natexlab{b}})Yu, Xing, Yuan, Hu, Li, Huang, Gao, Wong, Shan, and Tian]{ViewCrafter2024arXiv}
Wangbo Yu, Jinbo Xing, Li Yuan, Wenbo Hu, Xiaoyu Li, Zhipeng Huang, Xiangjun Gao, Tien-Tsin Wong, Ying Shan, and Yonghong Tian.
\newblock Viewcrafter: Taming video diffusion models for high-fidelity novel view synthesis.
\newblock \emph{arXiv preprint arXiv:2409.02048}, 2024{\natexlab{b}}.

\bibitem[Zhang et~al.(2018)Zhang, Isola, Efros, Shechtman, and Wang]{LPIPS2018CVPR}
Richard Zhang, Phillip Isola, Alexei~A Efros, Eli Shechtman, and Oliver Wang.
\newblock The unreasonable effectiveness of deep features as a perceptual metric.
\newblock In \emph{CVPR}, pages 586--595, 2018.

\bibitem[Zhang et~al.(2024)Zhang, Fei, Liu, Song, and Duan]{GGN2024NIPS}
Shengjun Zhang, Xin Fei, Fangfu Liu, Haixu Song, and Yueqi Duan.
\newblock Gaussian graph network: Learning efficient and generalizable gaussian representations from multi-view images.
\newblock \emph{NeurIPS}, 37:\penalty0 50361--50380, 2024.

\bibitem[Zhou et~al.(2018)Zhou, Tucker, Flynn, Fyffe, and Snavely]{Re10k2018arXiv}
Tinghui Zhou, Richard Tucker, John Flynn, Graham Fyffe, and Noah Snavely.
\newblock Stereo magnification: Learning view synthesis using multiplane images.
\newblock \emph{arXiv preprint arXiv:1805.09817}, 2018.

\end{thebibliography}


{
    \small
    

}
\end{document}